\documentclass{bmvc2k}
\usepackage{marvosym}

\def\blue#1{\textcolor{blue}{#1}}

\def\blue#1{\textcolor{blue}{#1}}

\title{BOTM: Echocardiography Segmentation \\via Bi-directional Optimal Token Matching}

\addauthor{Zhihua Liu}{zl208@leicester.ac.uk}{1}
\addauthor{Lei Tong}{lei.tong1@astrazeneca.com}{2}
\addauthor{Xilin He}{xilinhe@cuhk.edu.hk}{3}
\addauthor{Che Liu}{cl522@ic.ac.uk}{4}
\addauthor{Rossella Arcucci}{r.arcucci@imperial.ac.uk}{4}
\addauthor{Chen Jin}{chen.jin@astrazeneca.com}{2}
\addauthor{Huiyu Zhou}{hz143@leicester.ac.uk}{1,\blue{\text{\Letter}}}

\addinstitution{
 School of Computing and Mathematical Sciences, \\
 University of Leicester\\
 Leicester, UK 
}
\addinstitution{
 Centre for AI, AstraZeneca\\
 Cambridge, UK
}
\addinstitution{
 Information Engineering Department\\
 The Chinese University of Hong Kong\\
 Hong Kong SAR, China
}
\addinstitution{
 Data Science Institute\\
 Imperial College London\\
 London, UK \\
 \vspace{0.5cm}
 \blue{\text{\Letter}}: \footnotesize{Corresponding author}
}

\runninghead{Liu et al}{BOTM}

\usepackage{times}
\usepackage{soul}
\usepackage{url}
\usepackage{graphicx}
\usepackage{amsmath}
\usepackage{amsthm}
\usepackage{amssymb}
\usepackage{multirow}
\usepackage{booktabs}
\usepackage{algorithm}
\usepackage{algorithmic}
\usepackage{bm}
\usepackage[switch]{lineno}
\usepackage{float}
\usepackage{subfigure}
\usepackage{wrapfig}
\usepackage{caption}
\usepackage{adjustbox}
\usepackage{rotating}

\usepackage{float}
\usepackage{colortbl}

\def\eg{\emph{e.g.}}

\def\ie{\emph{i.e.}}

\definecolor{hl}{gray}{.9}

\def\blue#1{\textcolor{blue}{#1}}

\def\eg{\emph{e.g}\bmvaOneDot}

\definecolor{darkgray}{gray}{0.45}
\newcounter{rownumbers}
\renewcommand\rownum{{\color{darkgray} \stepcounter{rownumbers}\arabic{rownumbers}.}}

\DeclareMathOperator*{\argmin}{arg\,min}

\begin{document}

\maketitle

\begin{abstract}
Existed echocardiography segmentation methods often suffer from anatomical inconsistency challenge caused by shape variation, partial observation and region ambiguity with similar intensity across 2D echocardiographic sequences, resulting in false positive segmentation with anatomical defeated structures in challenging low signal-to-noise ratio conditions. To provide a strong anatomical guarantee across different echocardiographic frames, we propose a novel segmentation framework named \textbf{BOTM} (\textbf{B}i-directional \textbf{O}ptimal \textbf{T}oken \textbf{M}atching) that performs echocardiography segmentation and optimal anatomy transportation simultaneously. Given paired echocardiographic images, BOTM learns to match two sets of discrete image tokens by finding optimal correspondences from a novel anatomical transportation perspective. We further extend the token matching into a bi-directional cross-transport attention proxy to regulate the preserved anatomical consistency within the cardiac cyclic deformation in temporal domain. Extensive experimental results show that BOTM can generate stable and accurate segmentation outcomes (\eg $-1.917$ HD on CAMUS2H LV, $+1.9\%$ Dice on TED), and provide a better matching interpretation with anatomical consistency guarantee.
\end{abstract}

\section{Introduction}
\label{sec:intro}
Cardiac dysfunction stands as a primary cause for hospital admissions, representing a growing concern in global health~\cite{ziaeian2016epidemiology}. Measuring cardiac biomarkers deformation between end-diastolic (ED) and end-systolic (ES) volume is critical for assessing cardiac function and identifying patients eligible for life-prolonging therapies~\cite{loehr2008heart}. Echocardiography serves as a primary imaging modality for a range of medical professionals, including cardiologists, surgeons, oncologists, and emergency physicians, supporting diagnostic decisions, risk stratification, treatment planning, and surgical preparation due to its advantages of being low-cost, rapidly-acquired, radiation-free, and non-invasiveness~\cite{lang2015recommendations}. 
However, manual echocardiogrphy segmentation is time-consuming and highly dependent on the clinician's expertise, while also being subject to inter- and intra-observer variability~\cite{farsalinos2015head,lang2015recommendations}. Automated segmentation methods aim to overcome these challenges by accurately delineating cardiac anatomical structures from echocardiographic sequences or key frames between ED and ES. However, accurate and stable echocardiography segmentation remains a significant challenge due to speckle noise, anatomical variability, partial visibility, and region ambiguities (Fig.~\ref{fig:Challenges}\textbf{(a)}), resulting in disconnected boundaries, ambiguous localization and topological defects (Fig.~\ref{fig:Challenges}\textbf{(b)}).

\begin{figure*}[t!]
	\begin{center}
	\includegraphics[width=0.9\textwidth]{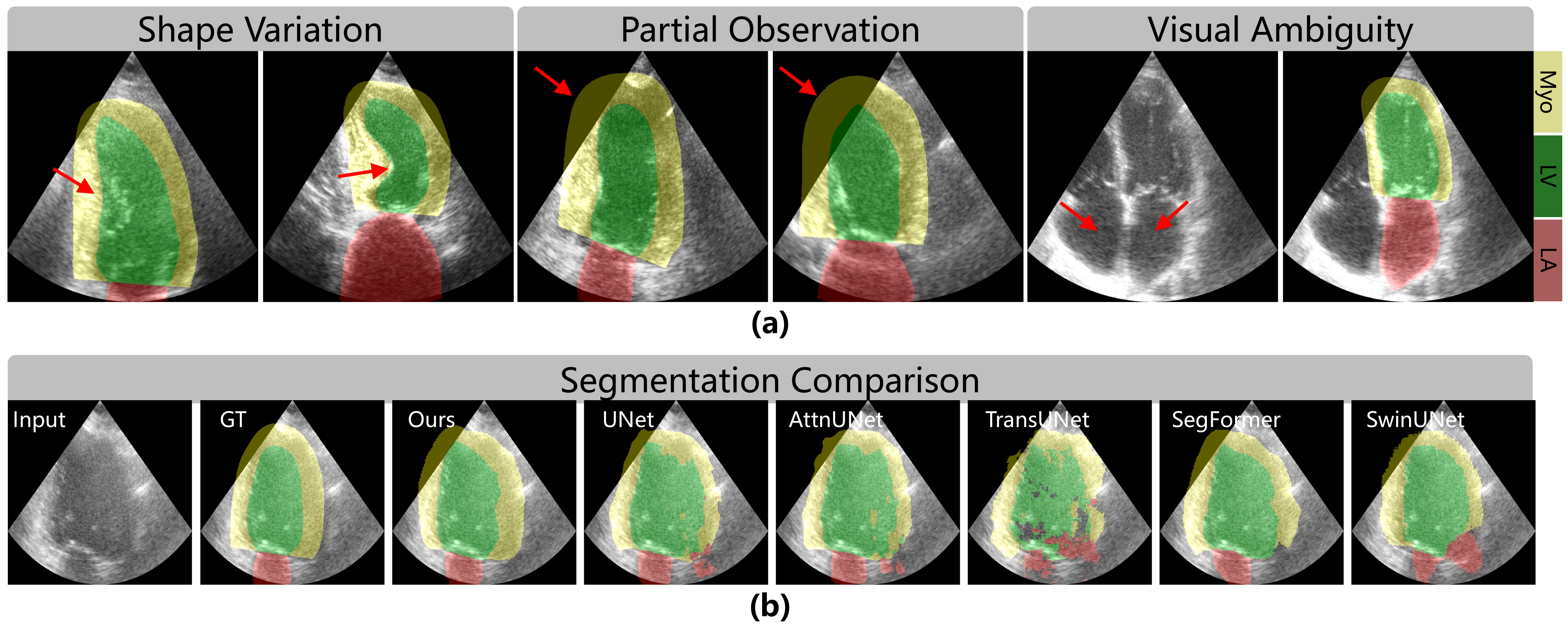}
        \end{center}
	\caption{Illustration of various echocardiography segmentation challenges and performance comparison:~\textbf{(a)}~Representative examples of echocardiography segmentation challenges, including shape variation across individuals, partial observations due to limited field-of-view, and visual ambiguity in regions with similar intensities. \textbf{(b)}~Qualitative comparison of segmentation performance. Our BOTM can achieve accurate segmentation, whereas others are affected by different types of noise, resulting in anatomical defeated mask.}
    \label{fig:Challenges}
\end{figure*}

\begin{figure*}[t!]
	\begin{center}
	\includegraphics[width=0.9\textwidth]{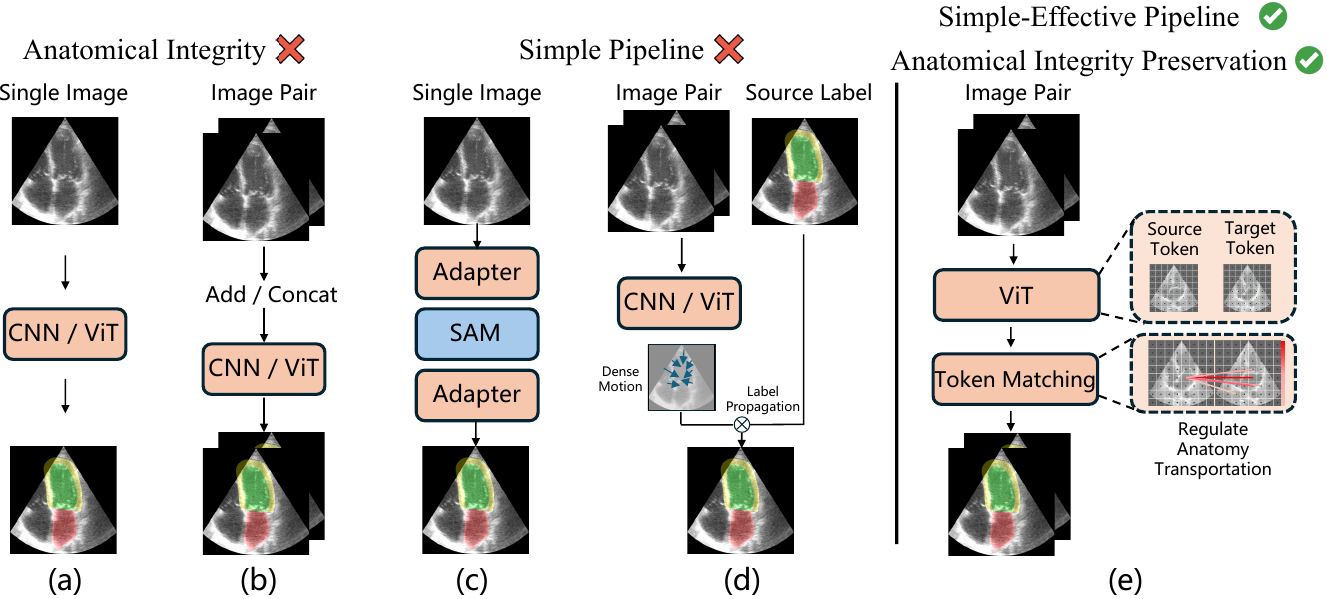}
        \end{center}
	\caption{Key comparison of echocardiography segmentation methods, including traditional end-to-end methods~\cite{wu2022fat,he2023h2former} with (a) single frame input and (b) image pair input; (c) SAM with echocardiography domain adapter including~\cite{lin2024beyond,gowda2024cc}; (d) motion tracking based segmentation methods including~\cite{kim2022diffusemorph,yang2024bidirectional} and (e) our token-matching based method, which regulates anatomical information transportation through a simple-yet-effective token matching pipeline for better semantical and anatomical structural coherent segmentation.}
    \label{fig:High-Level Comparison}
\end{figure*}

Early echocardiography segmentation methods are predominantly based on UNet~\cite{ronneberger2015u}, vision transformer~\cite{dosovitskiy2021an} and their variants~\cite{oktay2022attention,chen2021transunet, wu2022fat,xie2021segformer,cao2022swin, he2023h2former}. However, these models often overfit to individual frames, which may lead to false positive segmentation (Fig.~\ref{fig:High-Level Comparison}(a)). Some studies have attempted to incorporate temporal information across frames~\cite{ahn2021multi,dai2022cyclical,wei2020temporal}, but typically only through simple frame addition or concatenation. These models often produce anatomically inconsistent segmentation, as they lack mechanisms for preserving anatomical and structural integrity. (Fig.~\ref{fig:High-Level Comparison}(b)). Recent studies~\cite{lin2024beyond,gowda2024cc} have introduced domain-specific adapter modules on the top of Segment Anything Model (SAM)~\cite{kirillov2023segment} capitalizing on its fine-grained instance recognition capabilities. However, this type of approach adds additional computational complexity and highly depends on proper mask prompts, which can complicate the segmentation workflow (Fig.~\ref{fig:High-Level Comparison}(c)). Further improvements in cardiac segmentation have been achieved through jointly optimizing regularization strategies during training~\cite{painchaud2020cardiac,painchaud2022echocardiography,zheng2023gl,chen2023deep}. Yet, these approaches depend on pixel-wise constraints, such as full-resolution dense motion fields, which often result in redundancy due to their sensitivity to per-pixel modeling. In contrast, our proposed method operates anatomical consistency modeling on patch-level tokens, offering a more efficient representation that is less susceptible to pixel-level variations. (Fig.~\ref{fig:High-Level Comparison}(d)).

In this paper, we propose \textbf{BOTM}, a \textbf{B}i-directional \textbf{O}ptimal \textbf{T}oken \textbf{M}atching framework for echocardiography segmentation (Sec.~\ref{Sec:Methods}). \textit{Our motivation is driven by the clinical need to maintain anatomical consistency across frames through continuous segmentation, preserving anatomical details and ensuring that corresponding structures retain their identity over time.} Different from previous methods (Fig.~\ref{fig:High-Level Comparison}), our approach learns and enforces token-level anatomical consistency from a novel optimal transport (OT) perspective, without relying on complex pre- or post-processing (Fig.~\ref{fig:High-Level Comparison}(e)). 
In summary, our contributions are threefold: (1) We introduce a novel token matching framework named BOTM for challenging echocardiography segmentation. Our method incorporates jointly optimized anatomical consistency regularization, ensuring implicit anatomy preservation via optimal transport. (2) We extend the token matching into a bi-directional cross-transport attention proxy, where we can aggregate the anatomical priors from both spatial and temporal domains. (3) Extensive experiments demonstrate that BOTM can achieve competitive or superior segmentation with strong generalization and high interpretation across diverse scenarios.

\section{Methods}
\label{Sec:Methods}
\begin{figure*}[t!]
	\begin{center}
	\includegraphics[width=\textwidth]{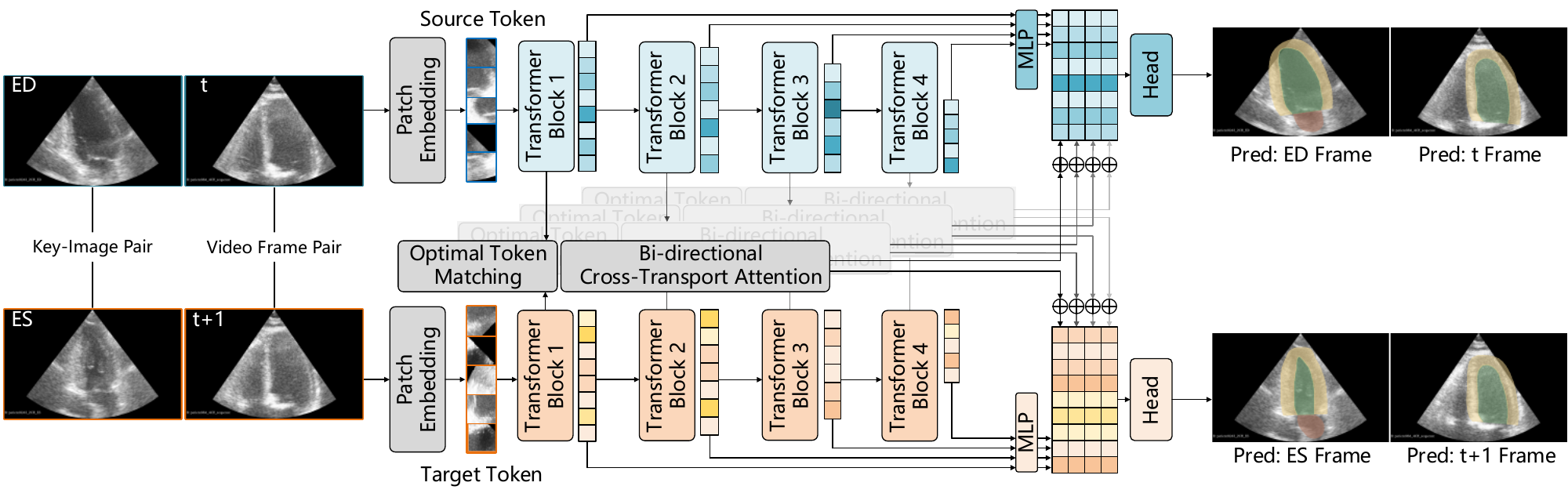}
        \end{center}
	\caption{Pipeline of the proposed \textbf{BOTM}: We first extract frame-dependent token embeddings using a vision transformer. At each embedding stage within the shared vision transformer encoder, BOTM learns optimal token matching correspondences via a bi-directional cross-transport attention mechanism. This serves as a proxy module to enforce implicit anatomical consistency across echocardiographic frames during the embedding process.}
    \label{fig:Pipeline}
\end{figure*}

Fig.~\ref{fig:Pipeline} shows the pipeline of our proposed BOTM. Given a pair of echocardiography images with source and target frames $\{I_{s}, I_{t}\}$, we utilize a shared vision transformer encoder to obtain the token embedding hierarchy $\boldsymbol{X}_{s} = \{X_{s}^{1},\cdots, X_{s}^{L}\},~\boldsymbol{X}_{t} = \{X_{t}^{1}, \cdots, X_{t}^{L}\}$. Within each encoding stage, a bi-directional cross-transport attention act as a proxy module (Fig.~\ref{fig:OTM_and_BCTA}(a)) aiming to solve the optimal transport problem by matching anatomically corresponded tokens (Fig.~\ref{fig:OTM_and_BCTA}(b)). Each component will be elaborated in detail as follows.

\subsection{Anatomical Consistency as Optimal Token Matching}
\label{Sec:Anatomical Consistency as Optimal Token Matching}
We formulate anatomical consistency estimation as learning an optimal transport (OT) map $\mathbf{T}^{\star}$, where each entry represents the dense matching scores between token embedding instances of $X_{s}^{l}\in \mathbb{R}^{h_{s}^{l}\times w_{s}^{l}\times d_{s}^{l}}$ and $X_{t}^{l} \in \mathbb{R}^{h_{t}^{l}\times w_{t}^{l}\times d_{s}^{l}}$ at each stage $l = \{1, 2, ..., L\}$. We first define the cost matrix $\mathbf{C} = 1 - \mathbf{M}$ as the subtraction of token similarity $\mathbf{M} = \frac{X_{s}^{l} \cdot X_{t}^{l}}{||X_{s}^{l}|| \cdot ||X_{t}^{l}||} \in \mathbb{R}^{h_{s}^{l}w_{s}^{l} \times h_{t}^{l}w_{t}^{l}}$. Then we solve the linear programming in the original optimal transport problem~\cite{cuturi2013sinkhorn} by equivalently minimizing the matching difference between the total token embedding instances:

\begin{equation}
\mathbf{T}_{l}^{\star} = \argmin_{\mathbf{T}\in \mathbb{R}^{h_{s}^{l}w_{s}^{l} \times h_{t}^{l}w_{t}^{l}}} \sum_{ij}\mathbf{T}_{ij}\mathbf{C}_{ij} + \epsilon H(\mathbf{T})
\label{Eqn: OT Dual}
\end{equation}

Eqn.~(\ref{Eqn: OT Dual}) is a strongly convex by 
resorting to the original OT with entropy $H(\mathbf{T})$ regularized by temprature $\epsilon = 0.1$. As illustrated in Fig.~\ref{fig:OTM_and_BCTA}(a), we obtain the optimal matching score map as a normalized exponential matrix using the iterative Sinkhorn algorithm~\cite{peyre2019computational}.

\subsection{Bi-directional Cross-Transport Attention}
\label{Sec:Bi-directional Token Mask Routing}
Given the token-level optimal transport plan $\mathbf{T}_{l}^{\star}$ from Eqn.(\ref{Eqn: OT Dual}), a na\"ive implementation of matching token embeddings between $\mathbf{X}_{s}^{l}$ and $\mathbf{X}_{t}^{l}$ is utilizing a partition method, such as Hungarian algorithm~\cite{munkres1957algorithms} as a post-processing step. \textit{This partition method operates under the assumption that each token holds identical significance, implying that the segmentation contribution of each token is uniformly distributed.} This assumption contradicts the learning process, where large area in background typically contribute \textbf{less} to the final segmentation compared to smaller but important structures such as the myocardium~\cite{oktay2017anatomically,You2023}. Thus, a dynamic token-level anatomical importance estimation should be adopted. Inspired by~\cite{sarlin2020superglue,weinzaepfel2021learning}, we reformulate this optimal token matching as a novel \textbf{B}i-directional \textbf{C}ross-\textbf{T}ransport \textbf{A}ttention proxy module (\textbf{BCTA}), where the transport attention $\mathbf{A}_{k}$ calculates the barycentric interpolated embedding in both forward and backward direction $\mathbf{A}_{k} = \mathbf{X}_{k}(\mathbf{X}_{\setminus k} \otimes \mathbf{T}_{l}^{\star}) / \sqrt{D}, \text{for}~k\in\{s, t\}$, where $D$ is the dimension of $\mathbf{X}$.

\begin{figure*}[t!]
	\begin{center}
	\includegraphics[width=\textwidth]{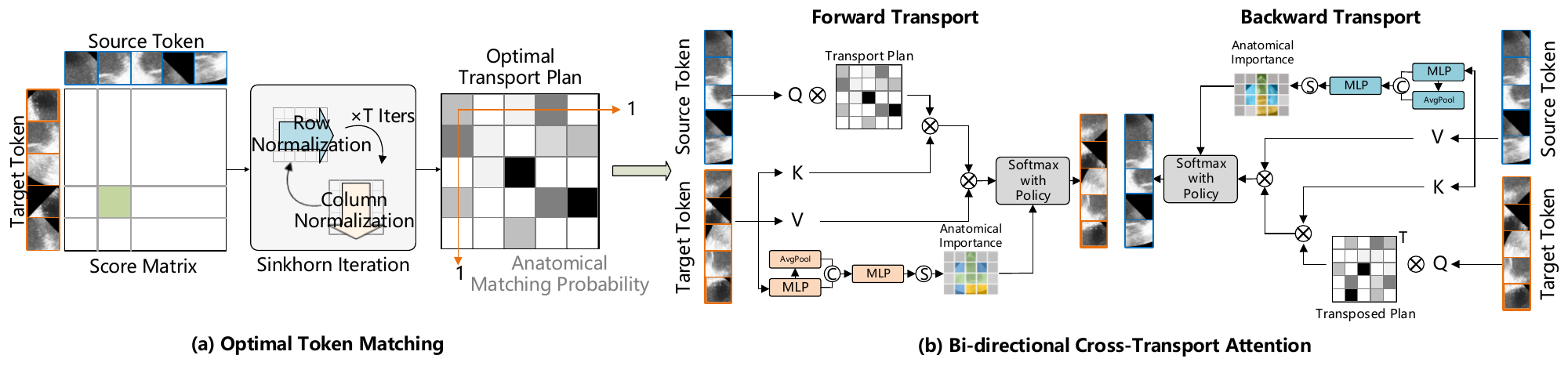}
        \end{center}
	\caption{(a) Given a pair of image token embeddings, we first compute an optimal transport plan using Sinkhorn iterations, where each entry represents a matching probability derived from cosine similarity between embedding instances. (b) We then refine the paired embeddings via a cross-attention mechanism that incorporates the previously computed transport plan. A learnable anatomical importance mask is applied to suppress regions with high matching probability but low anatomical relevance, such as background areas.}
    \label{fig:OTM_and_BCTA}
\end{figure*}

Different from standard cross-attention~\cite{chen2021crossvit,huang2023accurate,lin2023few}, our cross-transport attention updates token embedding by incorporating a hybrid policy derived from both anatomical transportation plan $\mathbf{T}_{l}^{\star}$ and anatomical importance $\mathbf{Z}_{k}^{l}$ in bi-direction. Following~\cite{yan2022sam}, we obtain the local anatomical saliency map $\mathbf{Z}^{l}_{k, local}$ through a light-weight MLP, and the global anatomical distribution map $\mathbf{Z}^{l}_{k, global}$ followed by an average pooling layer (Fig.~\ref{fig:OTM_and_BCTA}(b)).

\begin{equation}
    \begin{split}
        \mathbf{Z}_{k}^{l} = \textrm{Softmax}&(\textrm{MLP}(\textrm{Concat}[\mathbf{Z}_{k,local}^{l}, \mathbf{Z}_{k,global}^{l}])),\\
        \mathbf{Z}_{k,local}^{l} = \textrm{MLP}(\mathbf{X}_{k}^{l})&,~\mathbf{Z}_{k,global}^{l} = \textrm{AvgPool}(\textrm{MLP}(\mathbf{X}_{k}^{l})), k\in\{s, t\}
    \end{split}
\end{equation}

We then utilize the anatomical importance $\mathbf{Z}_{k}^{l}$ as a mask policy for calculating the cross-transport attention embedding for both source and target images in a bi-directional manner:

\begin{equation}
\begin{split}
    [\mathbf{\Tilde{X}}_{k}]_{ij} = \frac{\text{exp}([\mathbf{A}_{k}]_{ij})[\mathbf{P}_{k}^{l}]_{ij}}{\sum_{k=1}^{N}\text{exp}([\mathbf{A}_{k}]_{ik})[\mathbf{P}_{k}^{l}]_{ik}}, 
    \text{where}~[\mathbf{P}_{k}^{l}]_{ij} = \begin{cases}
        1,~\text{if}~i=j\\
        [\mathbf{Z}_{k}^{l}]_{ij},~\text{if}~i\neq j
    \end{cases},~1\leq~i,j\leq~N
\end{split}
\end{equation}

To this end, we undertake token matching by a bi-directional cross-transport attention proxy mechanism. $[\mathbf{P}_{k}^{l}]_{ij} = 1$ means the $j$-th token from $\mathbf{X}_{\setminus k}^{l}$ will contribute to the update of the $i$-th token in the corresponded $\mathbf{X}_{k}^{l}$. If $[\mathbf{P}_{k}^{l}]_{ij} = [\mathbf{Z}_{k}^{l}]_{ij} = 0$, the $j$-th token will not contribute to any other token. For generating semantic segmentation mask, we feed the transported token embeddings $\mathbf{\Tilde{X}}_{s}, \mathbf{\Tilde{X}}_{t}$ into a light-weight decoder with stacked four MLP layers used in~\cite{xie2021segformer}. The first two MLP unify the token embeddings from different stages into the same size. The last two MLP project the concatenated embedding to the final segmentation output.

\section{Experiments}
\label{Sec:Experiments}

\subsection{Implementation Details}
\noindent \textbf{Datasets.}~We evaluate BOTM on two public echocardiography datasets \textbf{CAMUS}~\cite{leclerc2019deep} and \textbf{TED}~\cite{painchaud2022echocardiography} with diverse settings. CAMUS is composed of 2D echocardiographic images of 500 patients collected on ED and ES frames. The official training set contains 450 patients and the test set contains 50 patients. We divide CAMUS into two subsets based on the view angle, \ie \textbf{CAMUS2CH} with apical two chamber view and \textbf{CAMUS4CH} with apical four chamber view. Both subsets contain the ground truth on left ventricle (LV), myocardium (Myo) and left atrium (LA). TED is the official refined dataset from CAMUS, composed of $2D+t$ echocardiography videos of 98 patients collected at apical four chamber view. Each frame contains the ground truth on LV and Myo. We randomly divide TED into 78 patients for training and 20 patients for testing.\\

\noindent \textbf{Training Settings.}~We implement BOTM in PyTorch, which is accelerated by a single NVIDIA A100 GPU. We employ the SGD optimizer with a learning rate of $1e-3$ and momentum $0.9$. All methods are trained with 500 epochs with a batch size of 8. We adopt the standard \textit{Dice} and \textit{Cross-Entropy} loss function with the same weight during the training process. All input images are resized to $672\times672$. We employ random horizontal flip ($p=0.5$) as data augmentation. Our code will be released upon acceptance.\\

\setcounter{rownumbers}{0}
\begin{table}[t]
\begin{center}
    \small
    \scriptsize
    \adjustbox{max width=.99\textwidth}{
    \begin{tabular}{rl  l  ccc  ccc}
    \toprule
    & \multirow{3}{*}{Model} & \multirow{3}{*}{Venue} & 
        \multicolumn{6}{c}{\textbf{CAMUS2CH}} \\
        &&&
         \multicolumn{2}{c}{Left Ventricle (LV)}  &
         \multicolumn{2}{c}{Left Atrium (LA)}  &
         \multicolumn{2}{c}{Myocardium (Myo)}  \\
        &&&
        mDice & mHD &
        mDice & mHD & 
        mDice & mHD \\
    \midrule
    \multicolumn{9}{l}{\emph{End-to-end segmentation networks}} \\
    
    \rownum & UNet~\cite{ronneberger2015u} & MICCAI15 & 0.914 & 8.248 & 0.883 & 17.425 & 0.892 & \textbf{7.192} \\
    \rownum & AttnUNet~\cite{oktay2022attention} & MIDL18 & 0.920 & 11.654 & 0.890 & 9.951 & 0.888 & 9.074 \\
    \rownum & SegFormer~\cite{xie2021segformer} & NeurIPS21 & 0.892	& 16.686 & 0.834 & 18.691 & 0.803 & 22.831 \\ 
    \rownum & SwinUNet~\cite{cao2022swin} & ECCVW22 & 0.910 & 7.821 & 0.890 & 8.777 & 0.872 & 10.112 \\
    \rownum & FAT-Net~\cite{wu2022fat}$^*$ & MIA22 & 0.936 & 9.200 & 0.916 & 12.050 & 0.872 & 15.930 \\
    \rownum & H2Former~\cite{he2023h2former}$^*$ & TMI23 & 0.934 & 9.600 & 0.910 & 11.920 & 0.873 & 16.600 \\
    \rownum & TransUNet~\cite{chen2021transunet} & MIA24 & 0.896 & 12.250 & 0.879 & 10.300 & 0.874 & 10.498 \\
    \multicolumn{9}{l}{\emph{Segmentation with anatomical motion tracking}} \\
    \rownum & DiffuseMorph~\cite{kim2022diffusemorph}$^{**}$ & ECCV22 & 0.858	& - & 0.879 & - & 0.874 & - \\  
    \rownum & GPTrack~\cite{yang2024bidirectional}$^{**}$ & NeurIPS24 & 0.886 & - & 0.891 & - & 0.804 & - \\
    \multicolumn{9}{l}{\emph{Segmentation with SAM adapters}} \\
    \rownum & SAMUS~\cite{lin2024beyond}$^*$ & MICCAI24 & 0.937 & 9.790 & 0.916 & 11.600 & 0.875 & 16.740 \\
    \rownum & CC-SAM~\cite{gowda2024cc}$^*$ & ECCV24 & \textbf{0.940} & 9.110 & \textbf{0.920} & 11.110 & 0.883 & 16.110 \\
    \cmidrule{2-9}
    \rownum & BOTM & - & 0.936 & \textbf{5.904} & 0.911 & \textbf{5.862} & \textbf{0.918} & 8.027 \\
    
    \bottomrule
    \end{tabular}
    }
    \end{center}
    \caption{
        Quantitative results comparison on CAMUS apical two chamber view (CAMUS2CH), \textbf{the most commonly used dataset and view setting for echocardiography segmentation.} Results with $^*$ are from CC-SAM~\cite{gowda2024cc} and $^{**}$ from GP-Track~\cite{yang2024bidirectional}.
    }
    \label{tab:CAMUS2CH}
\end{table}

\begin{figure}[!t]
\small
    \centering
    \begin{minipage}[b]{0.48\textwidth}
     \begin{center}     
    \resizebox{0.9\linewidth}{!}{
    \setcounter{rownumbers}{0}
    \begin{tabular}{rl  ccccc}
    \toprule
    & \multirow{3}{*}{Model} & 
        \multicolumn{5}{c}{\textbf{CAMUS4CH}} \\
        
        &&
        mDice & mHD &
        mIoU & Spe & 
        MAE\\
    \midrule    
    \rownum & UNet~\cite{ronneberger2015u}  & 0.893 & 10.776 & 0.869 & 0.993 & 0.015 \\
    \rownum & AttnUNet~\cite{oktay2022attention} & 0.894 & 11.646 & 0.868 & 0.992 & 0.014 \\
    \rownum & SegFormer~\cite{xie2021segformer} & 0.880 & 13.272 & 0.799 & 0.992 & 0.018 \\ 
    \rownum & SwinUNet~\cite{cao2022swin} & 0.881 & 10.649 & 0.858 & 0.988 & 0.022 \\
    \rownum & TransUNet~\cite{chen2021transunet} & 0.855 & 15.575 & 0.782 & 0.987 & 0.031 \\
    \cmidrule{2-7}
    \rownum & BOTM & \textbf{0.916} & \textbf{6.636} & \textbf{0.893} & \textbf{0.994} & \textbf{0.013} \\
    
    \bottomrule
    \end{tabular}
    }
    \vspace{0.2cm}
    \captionof{table}{Quantitative results comparison on CAMUS4CH. Results are averaged across all regions and all frames.}
    \label{tab:CAMUS4CH}
    \end{center}
    \end{minipage}
    \begin{minipage}[b]{0.48\textwidth}
    \begin{center}
    \resizebox{0.9\linewidth}{!}{
    \setcounter{rownumbers}{0}
    \begin{tabular}{rl  ccccc}
    \toprule
    & \multirow{3}{*}{Model} & 
        \multicolumn{5}{c}{\textbf{TED}} \\
        &&
        mDice & mHD &
        mIoU & Spe & 
        MAE\\
    \midrule    
    \rownum & UNet~\cite{ronneberger2015u}  & 0.901 & 11.316 & 0.844 & 0.986  & 0.022 \\
    \rownum & AttnUNet~\cite{oktay2022attention} & 0.904 & 9.570 & 0.858 & 0.987  & 0.021 \\
    \rownum & SegFormer~\cite{xie2021segformer} & 0.886 & 14.289 & 0.736 & 0.987  & 0.024 \\ 
    \rownum & SwinUNet~\cite{cao2022swin} & 0.872 & 13.149 & 0.792 & 0.984  & 0.030 \\
    \rownum & TransUNet~\cite{chen2021transunet} & 0.850 & 13.627 & 0.774 & 0.986  & 0.031 \\
    \cmidrule{2-7}
    \rownum & BOTM & \textbf{0.923} & \textbf{7.519} & \textbf{0.893} & \textbf{0.993} & \textbf{0.014} \\
    
    \bottomrule
    \end{tabular}
    }
    \vspace{0.2cm}
    \captionof{table}{Quantitative results comparison on TED. Results are averaged across all regions and all frames.}
        \label{tab:TED}
    \end{center}
    \end{minipage}
\end{figure}

\noindent \textbf{Baselines.}~We compare BOTM with long-standing UNet~\cite{ronneberger2015u} and AttnUNet~\cite{oktay2017anatomically} and state-of-the-art transformer-based medical image segmentation TransUNet~\cite{chen2021transunet}, SwinUNet~\cite{cao2022swin} and SegFormer~\cite{xie2021segformer}. We also compare results from recent attempts, including end-to-end segmentation networks FAT-Net~\cite{wu2022fat} and H2Former~\cite{he2023h2former}, segment anything model with echocardiography adapter SAMUS~\cite{lin2024beyond} and CC-SAM~\cite{gowda2024cc} and segmentation with motion propagation DiffuseMorph~\cite{kim2022diffusemorph} and GPTranck~\cite{yang2024bidirectional}. \\

\noindent \textbf{Metrics.}~We employ mean Dice (mDice) and mean Hausdorff (mHD) for quantitative evaluation~\cite{leclerc2019deep,painchaud2022echocardiography}. To provide deeper insights, we further introduce three additional metrics that are widely used in medical segmentation: (1) mean Intersection-over-Union (mIoU), which measures the overlap between prediction and ground truth. (2) Specificity (Spe), which refers to the percentage of pixels that are negative and are classified as such. (3) Mean Absolute Error (MAE), which measures the pixel-level error between prediction and ground truth.\\

\begin{figure*}[!ht]
    \begin{center}
    \includegraphics[width=0.85\textwidth]{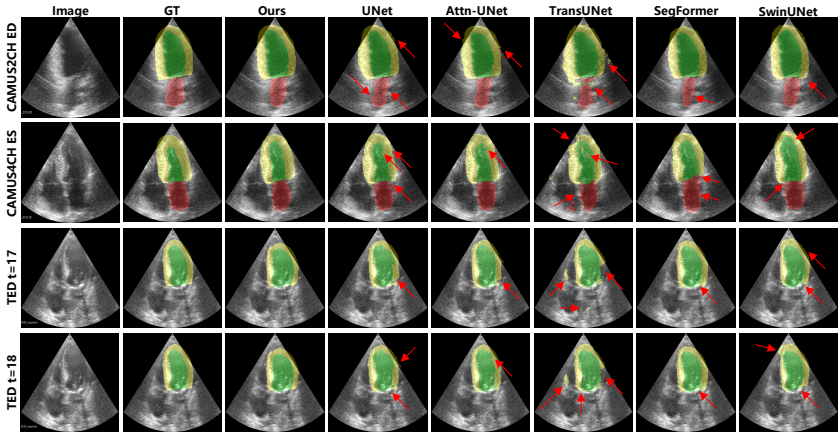}
    \end{center}
    \caption{\textbf{Qualitative comparison of segmentation results on CAMUS2CH, CAMUS4CH, and TED:} Our method produces accurate and anatomically meaningful segmentation masks, whereas baseline methods are adversely affected by various types of noise. This leads to anatomically compromised segmentations, characterized by disconnected boundaries, incomplete regions, and internal mask cavities (as indicated by red arrows). Please refer to the \textcolor{magenta}{supplementary material} for more results.}
    \label{fig:Image Qualitative Results}
\end{figure*}

\begin{table*}[!ht]
    \begin{center}
    \scriptsize
    \begin{tabular}{rcccccc}
        \hline
        & \multicolumn{3}{c}{RandomBlur} &\multicolumn{3}{c}{RandomGaussNoise}   \\
        \multirow{-2}{*}{Methods}&   10\%  &   30\%  &   50\%  &   10\%  &   30\%  & 50\% \\
        \hline
        UNet~\cite{ronneberger2015u}            & 0.902/\textbf{0.897} & 0.802/0.839 & 0.712/0.698 & 0.897/0.885 & 0.804/0.767 & 0.674/0.625\\
        TransUNet~\cite{chen2021transunet}            &  0.869/0.874 & 0.778/0.813 & 0.693/0.742 & 0.881/0.879 & 0.769/0.803 & 0.712/0.794 \\
        \textbf{BOTM(Ours)}     & \textbf{0.906}/0.892 & \textbf{0.895/0.887} & \textbf{0.862/0.858} & \textbf{0.900/0.907} & \textbf{0.873/0.887} & \textbf{0.832/0.841}\\
        \hline
    \end{tabular}
    \end{center}
    \caption{\textbf{Generalization study on CAMUS4CH:} We introduce test-time artifact noise using \textit{RandomBlur} and \textit{RandomGaussianNoise} to evaluate robustness. Reported results are averaged across all anatomical regions among all test samples. The left value corresponds to the end-diastolic (ED) frame and the right value to the end-systolic (ES) frame.}
\label{tab:GeneralizationCAMUS}
\end{table*}

\begin{table*}[!ht]
    \small
\begin{center}
    \scriptsize
    \begin{tabular}{rcccc}
        \hline
       & \multicolumn{4}{c}{RandomFrameDropout} \\
        \multirow{-2}{*}{Methods}&   10\%  &   30\%  &   50\%  &   70\%\\
        \hline
        UNet~\cite{ronneberger2015u}            & 0.901 & 0.877 & 0.849 & 0.810\\
        TransUNet~\cite{chen2021transunet}            & 0.869 & 0.833 & 0.802 & 0.734 \\
        \textbf{BOTM(Ours)}     & \textbf{0.912} & \textbf{0.893} & \textbf{0.875} & \textbf{0.851} \\
        \hline
    \end{tabular}
    \end{center}
    \caption{\textbf{Generalization study on TED:} we randomly drop training frames using \textit{RandomFrameDropout} to evaluate robustness. Results are averaged across all anatomical regions and all frames in the test videos.}
\label{tab:GeneralizationTED}
\end{table*}

\subsection{Evaluation on Echocardiography Segmentation}
\label{Sec: Evaluation on Echocardiography Segmentation}

\noindent \textbf{Learning Ability.}~We conduct extensive experiments to evaluate the learning capability of BOTM across CAMUS2CH, CAMUS4CH, and TED. As shown in Table~\ref{tab:CAMUS2CH}, BOTM achieves state-of-the-art segmentation performance across various metrics, including a substantial reduction in mHD ($-1.917$ for LV, $-2.915$ for LA), while maintaining competitive results in terms of mDice.This performance is notable even when compared to SAMUS~\cite{lin2024beyond} and CC-SAM~\cite{gowda2024cc}, which incorporate the powerful fine-grained segmentation capabilities of SAM~\cite{kirillov2023segment} with customized adapters and prompts for echocardiography segmentation. Together with superior performance on the apical four chamber view segmentation (Table~\ref{tab:CAMUS4CH}) and video segmentation tasks (Table~\ref{tab:TED}), these results collectively suggest that, by simply enforce the anatomical consistency through token matching, BOTM demonstrates strong learning capacity and robustness under diverse imaging conditions for effective echocardiography segmentation, without complex ad-hoc modules or manual visual prompts. For detailed sub-class segmentation results, please refer to the table in \textcolor{magenta}{supplementary material}.\\

\noindent \textbf{Generalization Capability.}~We evaluated the robustness of BOTM under challenging conditions by manually introducing severe artifacts using \textit{RandomBlur} and \textit{RandomGaussianNoise} to the CAMUS4CH dataset. Additionally, we assessed segmentation stability under limited training data by applying \textit{RandomFrameDropout} to the TED video dataset. As shown in Table~\ref{tab:GeneralizationCAMUS} and Table~\ref{tab:GeneralizationTED}, BOTM consistently maintained stable performance, exhibiting a lower performance drop ratio compared to UNet and TransUNet. These results highlight that, by incorporating matching-based anatomical consistency, BOTM outperforms other state-of-the-art methods and demonstrates superior generalization capability under varying imaging conditions and data constraints.\\

\begin{figure*}[!t]
	\centering
	\includegraphics[width=0.9\textwidth]{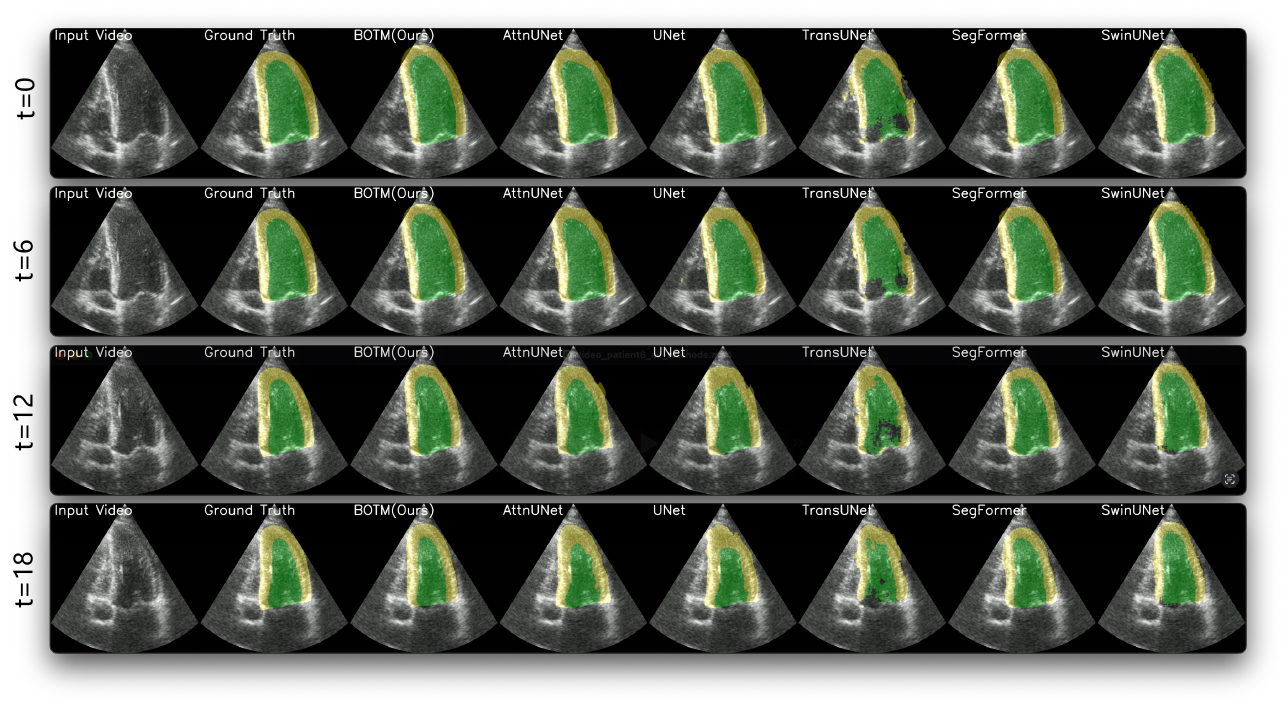}
        \vspace{0.1cm}
	\caption{\textbf{Qualitative comparison of echocardiography video segmentation results on TED:} Our method produces accurate, stable, and temporally consistent segmentation results across extended frame sequences. In contrast, baseline methods are adversely affected by noise within individual frames or accumulated over time, leading to unstable performance.}
    \label{fig:Video Qualitative Results}
\end{figure*}

\begin{figure*}[!t]
	\centering
	\includegraphics[width=\textwidth]{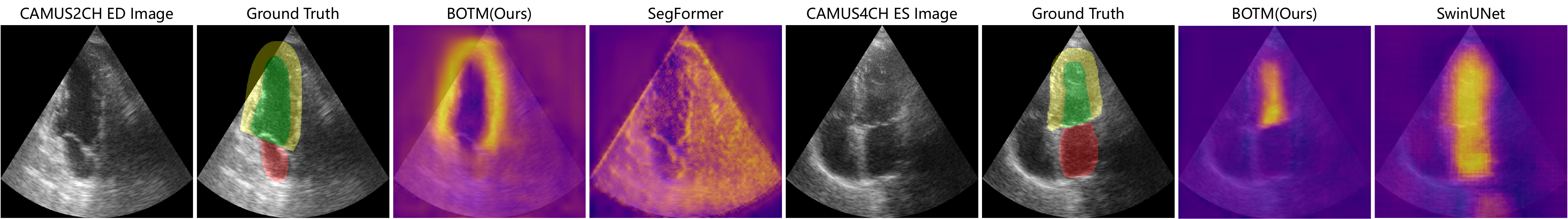}
        \vspace{0.1cm}
	\caption{\textbf{Qualitative comparison of segmentation uncertainty:} We visualize segmentation uncertainty for the myocardium (yellow) on CAMUS2CH (left) and the left ventricle (green) on CAMUS4CH (right). By incorporating token matching to enforce anatomical consistency, the proposed BOTM effectively reduces segmentation uncertainty, leading to more coherent and reliable mask boundary delineation.}
    \label{fig:Uncertainty}
\end{figure*}

\noindent \textbf{Qualitative Comparison.}~As shown in Fig.~\ref{fig:Image Qualitative Results} and \ref{fig:Video Qualitative Results}, BOTM demonstrates the ability to accurately locate and segment cardiac structures even under challenging conditions, while baseline methods fail to generate anatomically plausible segmentation with disconnect boundary, incomplete structure and mask cavity (red arrows). We further visualize the mask soft logits in Fig.~\ref{fig:Uncertainty}, illustrating that BOTM can effectively minimize the segmentation uncertainty. Leveraging cross-transport attention, BOTM effectively provides higher interpretation even when boundaries are blurred in severe narrowed ventricles. For additional qualitative comparison and details in zoomed-in patches, please refer to the \textcolor{magenta}{supplementary material}.

\subsection{Ablation Study}
\label{Sec:Ablation Study}

\begin{table*}[t]
\begin{center}
\scriptsize
\begin{tabular}{lcccccc}
 \hline 
 & \multicolumn{3}{c}{CAMUS2CH} &\multicolumn{3}{c}{TED}   \\
 \multirow{-2}{*}{Settings}&   mDice~$\uparrow$  &   mHD~$\downarrow$  &   mIoU~$\uparrow$  &   mDice~$\uparrow$  &   mHD~$\downarrow$  &  mIoU~$\uparrow$ \\
 \hline
(\#1) Base           & 0.874/0.886 & 14.161/12.668 & 0.802/0.779 & 0.891 & 13.181 & 0.810 \\
(\#2) Base + ADD          & 0.882/0.871 & 11.726/12.124 & 0.836/0.825 & 0.895 & 11.967 & 0.830 \\
(\#3) Base + OT        & 0.884/0.898 & 10.592/9.163 & 0.847/0.832 & 0.909 & 9.748 & 0.881 \\ 
(\#4) Base + OT + BCTA  & \textbf{0.908/0.934} & \textbf{7.334/5.861} & \textbf{0.887/0.884} & \textbf{0.923} & \textbf{7.519} & \textbf{0.893} \\
 \hline
\end{tabular}
\end{center}
\caption{Ablation study for BOTM key components on both CAMUS2CH and TED dataset. The left value corresponds to the end-diastolic (ED) frame and the right value to the end-systolic (ES) frame.}
\label{tab:Ablation}
\end{table*}

\noindent \textbf{Effectiveness of Paired Segmentation.}~We investigate the importance of segmentation with paired images. From Table~\ref{tab:Ablation}, we observe that by adding paired image features together (ADD) during encoding stage, $\#2$ slightly outperforms single-frame segmentation in $\#1$, indicating that cross-frame anatomical context is valuable for accurate segmentation.\\

\noindent \textbf{Effectiveness of Optimal Anatomical Transport.}~We investigate the contribution of anatomical consistency through OT, where we generate the updated token embedding by element-wise multiplication with the anatomical transport plan map. We observe that $\#3$ enhances performance compared to $\#2$, increasing around $2\%$ in the mDice. These improvements suggest that introducing anatomical consistency via optimal transport provides an implicit anatomical guarantee, enabling BOTM to accurately preserve shape continuity across frames.\\

\noindent \textbf{Effectiveness of Bi-Directional Transport Cross Attention Proxy.}~We test the formulation of the token matching proxy as bi-directional cross-transport attention ($\#4$). In Table~\ref{tab:Ablation}, $\#4$ outperforms $\#1\sim 3$. Furthermore, BOTM achieves superior results across diverse settings, including keyframe and video segmentation. These results demonstrate that BOTM is a simple-yet-effective and robust architecture, which can advance future research in echocardiography analysis and clinical index estimation.

\section{Conclusion}
 Accurate and stable segmentation in echocardiography remains a significant challenge due to various types of imaging noise. In this paper, we propose \textbf{BOTM}, a novel token matching framework for echocardiography segmentation. BOTM is motivated by the clinical need to maintain anatomical consistency across frames through continuous segmentation. To preserve anatomical details and ensure that corresponding structures retain their identity over time, BOTM introduces an implicit mechanism for enforcing anatomical consistency by formulating it as an optimal transport problem. This is further extended through a bi-directional cross-transport attention proxy, which dynamically constructs cross-attention guided by optimal transport plans and a learnable anatomical importance prior. Extensive experimental results demonstrate that BOTM achieves superior and stable segmentation performance, exhibiting strong learning capability, generalization, and interpretability across diverse settings.

 \section*{Acknowledgment}
 Most experiments were performed using the Sulis Tier 2 HPC platform hosted by the Scientific Computing Research Technology Platform at the University of Warwick. Sulis is funded by EPSRC Grant EP/T022108/1 and the HPC Midlands+ consortium.

\bibliography{bmvc_final}

\begin{thebibliography}{36}
\providecommand{\natexlab}[1]{#1}
\providecommand{\url}[1]{\texttt{#1}}
\expandafter\ifx\csname urlstyle\endcsname\relax
  \providecommand{\doi}[1]{doi: #1}\else
  \providecommand{\doi}{doi: \begingroup \urlstyle{rm}\Url}\fi

\bibitem[Ahn et~al.(2021)Ahn, Ta, Thorn, Langdon, Sinusas, and Duncan]{ahn2021multi}
Shawn~S Ahn, Kevinminh Ta, Stephanie Thorn, Jonathan Langdon, Albert~J Sinusas, and James~S Duncan.
\newblock Multi-frame attention network for left ventricle segmentation in 3d echocardiography.
\newblock In \emph{MICCAI}, pages 348--357. Springer, 2021.

\bibitem[Cao et~al.(2022)Cao, Wang, Chen, Jiang, Zhang, Tian, and Wang]{cao2022swin}
Hu~Cao, Yueyue Wang, Joy Chen, Dongsheng Jiang, Xiaopeng Zhang, Qi~Tian, and Manning Wang.
\newblock Swin-unet: Unet-like pure transformer for medical image segmentation.
\newblock In \emph{ECCV}, pages 205--218. Springer, 2022.

\bibitem[Chen et~al.(2021)Chen, Fan, and Panda]{chen2021crossvit}
Chun-Fu~Richard Chen, Quanfu Fan, and Rameswar Panda.
\newblock Crossvit: Cross-attention multi-scale vision transformer for image classification.
\newblock In \emph{IEEE ICCV}, pages 357--366, 2021.

\bibitem[Chen et~al.(2023)Chen, Chen, Kong, Zhang, Zheng, Sun, Zhang, and Liao]{chen2023deep}
Fang Chen, Lingyu Chen, Wentao Kong, Weijing Zhang, Pengfei Zheng, Liang Sun, Daoqiang Zhang, and Hongen Liao.
\newblock Deep semi-supervised ultrasound image segmentation by using a shadow aware network with boundary refinement.
\newblock \emph{IEEE TMI}, 2023.

\bibitem[Chen et~al.(2024)Chen, Mei, Li, Lu, Yu, Wei, Luo, Xie, Adeli, Wang, Lungren, Zhang, Xing, Lu, Yuille, and Zhou]{chen2021transunet}
Jieneng Chen, Jieru Mei, Xianhang Li, Yongyi Lu, Qihang Yu, Qingyue Wei, Xiangde Luo, Yutong Xie, Ehsan Adeli, Yan Wang, Matthew~P. Lungren, Shaoting Zhang, Lei Xing, Le~Lu, Alan Yuille, and Yuyin Zhou.
\newblock Transunet: Rethinking the u-net architecture design for medical image segmentation through the lens of transformers.
\newblock \emph{Medical Image Analysis}, 97:\penalty0 103280, 2024.
\newblock ISSN 1361-8415.
\newblock \doi{https://doi.org/10.1016/j.media.2024.103280}.

\bibitem[Cuturi(2013)]{cuturi2013sinkhorn}
Marco Cuturi.
\newblock Sinkhorn distances: Lightspeed computation of optimal transport.
\newblock \emph{NeurIPS}, 26, 2013.

\bibitem[Dai et~al.(2022)Dai, Li, Ding, and Cheng]{dai2022cyclical}
Weihang Dai, Xiaomeng Li, Xinpeng Ding, and Kwang-Ting Cheng.
\newblock Cyclical self-supervision for semi-supervised ejection fraction prediction from echocardiogram videos.
\newblock \emph{IEEE TMI}, 2022.

\bibitem[Dosovitskiy et~al.(2021)Dosovitskiy, Beyer, Kolesnikov, Weissenborn, Zhai, Unterthiner, Dehghani, Minderer, Heigold, Gelly, Uszkoreit, and Houlsby]{dosovitskiy2021an}
Alexey Dosovitskiy, Lucas Beyer, Alexander Kolesnikov, Dirk Weissenborn, Xiaohua Zhai, Thomas Unterthiner, Mostafa Dehghani, Matthias Minderer, Georg Heigold, Sylvain Gelly, Jakob Uszkoreit, and Neil Houlsby.
\newblock An image is worth 16x16 words: Transformers for image recognition at scale.
\newblock In \emph{International Conference on Learning Representations}, 2021.
\newblock URL \url{https://openreview.net/forum?id=YicbFdNTTy}.

\bibitem[Farsalinos et~al.(2015)Farsalinos, Daraban, {\"U}nl{\"u}, Thomas, Badano, and Voigt]{farsalinos2015head}
Konstantinos~E Farsalinos, Ana~M Daraban, Serkan {\"U}nl{\"u}, James~D Thomas, Luigi~P Badano, and Jens-Uwe Voigt.
\newblock Head-to-head comparison of global longitudinal strain measurements among nine different vendors: the eacvi/ase inter-vendor comparison study.
\newblock \emph{Journal of the American Society of Echocardiography}, 28\penalty0 (10):\penalty0 1171--1181, 2015.

\bibitem[Gowda and Clifton(2024)]{gowda2024cc}
Shreyank~N Gowda and David~A Clifton.
\newblock Cc-sam: Sam with cross-feature attention and context for ultrasound image segmentation.
\newblock In \emph{European Conference on Computer Vision}, pages 108--124. Springer, 2024.

\bibitem[He et~al.(2023)He, Wang, Li, Du, Xia, and Fu]{he2023h2former}
Along He, Kai Wang, Tao Li, Chengkun Du, Shuang Xia, and Huazhu Fu.
\newblock H2former: An efficient hierarchical hybrid transformer for medical image segmentation.
\newblock \emph{IEEE Transactions on Medical Imaging}, 42\penalty0 (9):\penalty0 2763--2775, 2023.

\bibitem[Huang et~al.(2023)Huang, Li, Mei, Zhang, Chen, Dong, Dong, Liu, and Lyu]{huang2023accurate}
Shoujin Huang, Jingyu Li, Lifeng Mei, Tan Zhang, Ziran Chen, Yu~Dong, Linzheng Dong, Shaojun Liu, and Mengye Lyu.
\newblock Accurate multi-contrast mri super-resolution via a dual cross-attention transformer network.
\newblock In \emph{MICCAI}, pages 313--322. Springer, 2023.

\bibitem[Kim et~al.(2022)Kim, Han, and Ye]{kim2022diffusemorph}
Boah Kim, Inhwa Han, and Jong~Chul Ye.
\newblock Diffusemorph: Unsupervised deformable image registration using diffusion model.
\newblock In \emph{European conference on computer vision}, pages 347--364. Springer, 2022.

\bibitem[Kirillov et~al.(2023)Kirillov, Mintun, Ravi, Mao, Rolland, Gustafson, Xiao, Whitehead, Berg, Lo, et~al.]{kirillov2023segment}
Alexander Kirillov, Eric Mintun, Nikhila Ravi, Hanzi Mao, Chloe Rolland, Laura Gustafson, Tete Xiao, Spencer Whitehead, Alexander~C Berg, Wan-Yen Lo, et~al.
\newblock Segment anything.
\newblock In \emph{Proceedings of the IEEE/CVF international conference on computer vision}, pages 4015--4026, 2023.

\bibitem[Lang et~al.(2015)Lang, Badano, Mor-Avi, Afilalo, Armstrong, Ernande, Flachskampf, Foster, Goldstein, Kuznetsova, et~al.]{lang2015recommendations}
Roberto~M Lang, Luigi~P Badano, Victor Mor-Avi, Jonathan Afilalo, Anderson Armstrong, Laura Ernande, Frank~A Flachskampf, Elyse Foster, Steven~A Goldstein, Tatiana Kuznetsova, et~al.
\newblock Recommendations for cardiac chamber quantification by echocardiography in adults: an update from the american society of echocardiography and the european association of cardiovascular imaging.
\newblock \emph{European Heart Journal-Cardiovascular Imaging}, 16\penalty0 (3):\penalty0 233--271, 2015.

\bibitem[Leclerc et~al.(2019)Leclerc, Smistad, Pedrosa, {\O}stvik, Cervenansky, Espinosa, Espeland, Berg, Jodoin, Grenier, et~al.]{leclerc2019deep}
Sarah Leclerc, Erik Smistad, Joao Pedrosa, Andreas {\O}stvik, Frederic Cervenansky, Florian Espinosa, Torvald Espeland, Erik Andreas~Rye Berg, Pierre-Marc Jodoin, Thomas Grenier, et~al.
\newblock Deep learning for segmentation using an open large-scale dataset in 2d echocardiography.
\newblock \emph{IEEE TMI}, 38\penalty0 (9):\penalty0 2198--2210, 2019.

\bibitem[Lin et~al.(2024)Lin, Xiang, Yu, and Yan]{lin2024beyond}
Xian Lin, Yangyang Xiang, Li~Yu, and Zengqiang Yan.
\newblock Beyond adapting sam: Towards end-to-end ultrasound image segmentation via auto prompting.
\newblock In \emph{International Conference on Medical Image Computing and Computer-Assisted Intervention}, pages 24--34. Springer, 2024.

\bibitem[Lin et~al.(2023)Lin, Chen, Cheng, and Chen]{lin2023few}
Yi~Lin, Yufan Chen, Kwang-Ting Cheng, and Hao Chen.
\newblock Few shot medical image segmentation with cross attention transformer.
\newblock In \emph{MICCAI}, pages 233--243. Springer, 2023.

\bibitem[Loehr et~al.(2008)Loehr, Rosamond, Chang, Folsom, and Chambless]{loehr2008heart}
Laura~R Loehr, Wayne~D Rosamond, Patricia~P Chang, Aaron~R Folsom, and Lloyd~E Chambless.
\newblock Heart failure incidence and survival (from the atherosclerosis risk in communities study).
\newblock \emph{The American journal of cardiology}, 101\penalty0 (7):\penalty0 1016--1022, 2008.

\bibitem[Munkres(1957)]{munkres1957algorithms}
James Munkres.
\newblock Algorithms for the assignment and transportation problems.
\newblock \emph{Journal of the society for industrial and applied mathematics}, 5\penalty0 (1):\penalty0 32--38, 1957.

\bibitem[Oktay et~al.(2017)Oktay, Ferrante, Kamnitsas, Heinrich, Bai, Caballero, Cook, De~Marvao, Dawes, O‘Regan, et~al.]{oktay2017anatomically}
Ozan Oktay, Enzo Ferrante, Konstantinos Kamnitsas, Mattias Heinrich, Wenjia Bai, Jose Caballero, Stuart~A Cook, Antonio De~Marvao, Timothy Dawes, Declan~P O‘Regan, et~al.
\newblock Anatomically constrained neural networks (acnns): application to cardiac image enhancement and segmentation.
\newblock \emph{IEEE TMI}, 37\penalty0 (2):\penalty0 384--395, 2017.

\bibitem[Oktay et~al.(2022)Oktay, Schlemper, Le~Folgoc, Lee, Heinrich, Misawa, Mori, McDonagh, Hammerla, Kainz, et~al.]{oktay2022attention}
Ozan Oktay, Jo~Schlemper, Loic Le~Folgoc, Matthew Lee, Mattias Heinrich, Kazunari Misawa, Kensaku Mori, Steven McDonagh, Nils~Y Hammerla, Bernhard Kainz, et~al.
\newblock Attention u-net: Learning where to look for the pancreas.
\newblock In \emph{Medical Imaging with Deep Learning}, 2022.

\bibitem[Painchaud et~al.(2020)Painchaud, Skandarani, Judge, Bernard, Lalande, and Jodoin]{painchaud2020cardiac}
Nathan Painchaud, Youssef Skandarani, Thierry Judge, Olivier Bernard, Alain Lalande, and Pierre-Marc Jodoin.
\newblock Cardiac segmentation with strong anatomical guarantees.
\newblock \emph{IEEE TMI}, 39\penalty0 (11):\penalty0 3703--3713, 2020.

\bibitem[Painchaud et~al.(2022)Painchaud, Duchateau, Bernard, and Jodoin]{painchaud2022echocardiography}
Nathan Painchaud, Nicolas Duchateau, Olivier Bernard, and Pierre-Marc Jodoin.
\newblock Echocardiography segmentation with enforced temporal consistency.
\newblock \emph{IEEE TMI}, 41\penalty0 (10):\penalty0 2867--2878, 2022.

\bibitem[Peyr{\'e} et~al.(2019)Peyr{\'e}, Cuturi, et~al.]{peyre2019computational}
Gabriel Peyr{\'e}, Marco Cuturi, et~al.
\newblock Computational optimal transport: With applications to data science.
\newblock \emph{Foundations and Trends{\textregistered} in Machine Learning}, 11\penalty0 (5-6):\penalty0 355--607, 2019.

\bibitem[Ronneberger et~al.(2015)Ronneberger, Fischer, and Brox]{ronneberger2015u}
Olaf Ronneberger, Philipp Fischer, and Thomas Brox.
\newblock U-net: Convolutional networks for biomedical image segmentation.
\newblock In \emph{MICCAI}, pages 234--241. Springer, 2015.

\bibitem[Sarlin et~al.(2020)Sarlin, DeTone, Malisiewicz, and Rabinovich]{sarlin2020superglue}
Paul-Edouard Sarlin, Daniel DeTone, Tomasz Malisiewicz, and Andrew Rabinovich.
\newblock Superglue: Learning feature matching with graph neural networks.
\newblock In \emph{IEEE CVPR}, pages 4938--4947, 2020.

\bibitem[Wei et~al.(2020)Wei, Cao, Cao, Zhou, Xue, Ni, and Li]{wei2020temporal}
Hongrong Wei, Heng Cao, Yiqin Cao, Yongjin Zhou, Wufeng Xue, Dong Ni, and Shuo Li.
\newblock Temporal-consistent segmentation of echocardiography with co-learning from appearance and shape.
\newblock In \emph{MICCAI}, pages 623--632. Springer, 2020.

\bibitem[Weinzaepfel et~al.(2021)Weinzaepfel, Lucas, Larlus, and Kalantidis]{weinzaepfel2021learning}
Philippe Weinzaepfel, Thomas Lucas, Diane Larlus, and Yannis Kalantidis.
\newblock Learning super-features for image retrieval.
\newblock In \emph{International Conference on Learning Representations}, 2021.

\bibitem[Wu et~al.(2022)Wu, Chen, Chen, Wang, Lei, and Wen]{wu2022fat}
Huisi Wu, Shihuai Chen, Guilian Chen, Wei Wang, Baiying Lei, and Zhenkun Wen.
\newblock Fat-net: Feature adaptive transformers for automated skin lesion segmentation.
\newblock \emph{Medical image analysis}, 76:\penalty0 102327, 2022.

\bibitem[Xie et~al.(2021)Xie, Wang, Yu, Anandkumar, Alvarez, and Luo]{xie2021segformer}
Enze Xie, Wenhai Wang, Zhiding Yu, Anima Anandkumar, Jose~M Alvarez, and Ping Luo.
\newblock Segformer: Simple and efficient design for semantic segmentation with transformers.
\newblock \emph{NeurIPS}, 34:\penalty0 12077--12090, 2021.

\bibitem[Yan et~al.(2022)Yan, Cai, Jin, Miao, Guo, Harrison, Tang, Xiao, Lu, and Lu]{yan2022sam}
Ke~Yan, Jinzheng Cai, Dakai Jin, Shun Miao, Dazhou Guo, Adam~P Harrison, Youbao Tang, Jing Xiao, Jingjing Lu, and Le~Lu.
\newblock Sam: Self-supervised learning of pixel-wise anatomical embeddings in radiological images.
\newblock \emph{IEEE TMI}, 41\penalty0 (10):\penalty0 2658--2669, 2022.

\bibitem[Yang et~al.(2024)Yang, Lin, Pu, and Li]{yang2024bidirectional}
Jiewen Yang, Yiqun Lin, Bin Pu, and Xiaomeng Li.
\newblock Bidirectional recurrence for cardiac motion tracking with gaussian process latent coding.
\newblock \emph{Advances in Neural Information Processing Systems}, 37:\penalty0 34800--34823, 2024.

\bibitem[You et~al.(2023)You, Dai, Min, Staib, Sekhon, and Duncan]{You2023}
Chenyu You, Weicheng Dai, Yifei Min, Lawrence Staib, Jas Sekhon, and James~S. Duncan.
\newblock Action++: Improving semi-supervised medical image segmentation with adaptive anatomical contrast.
\newblock In \emph{MICCAI}, pages 78--88. Springer, 2023.

\bibitem[Zheng et~al.(2023)Zheng, Yang, Ding, Xu, and Li]{zheng2023gl}
Ziyang Zheng, Jiewen Yang, Xinpeng Ding, Xiaowei Xu, and Xiaomeng Li.
\newblock Gl-fusion: Global-local fusion network for multi-view echocardiogram video segmentation.
\newblock In \emph{MICCAI}, pages 78--88. Springer, 2023.

\bibitem[Ziaeian and Fonarow(2016)]{ziaeian2016epidemiology}
Boback Ziaeian and Gregg~C Fonarow.
\newblock Epidemiology and aetiology of heart failure.
\newblock \emph{Nature Reviews Cardiology}, 13\penalty0 (6):\penalty0 368--378, 2016.

\end{thebibliography}

\newpage
\section{Supplementary}

\begin{figure*}[htbp]
	\begin{center}
	\includegraphics[width=0.9\textwidth]{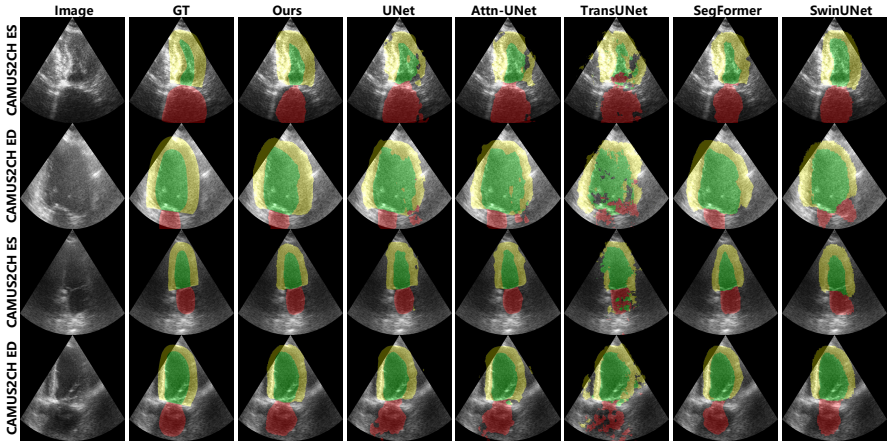}
        \end{center}
	\caption{Additional aualitative results on CAMUS apical two chamber view.}
    \label{fig:Challenges}
\end{figure*}

\begin{figure*}[htbp]
	\begin{center}
	\includegraphics[width=0.9\textwidth]{img/CAMUS4CHSupplementary.pdf}
        \end{center}
	\caption{Additional aualitative results on CAMUS apical four chamber view.}
    \label{fig:Challenges}
\end{figure*}

\newpage
\begin{figure*}[htbp]
	\begin{center}
	\includegraphics[width=0.9\textwidth]{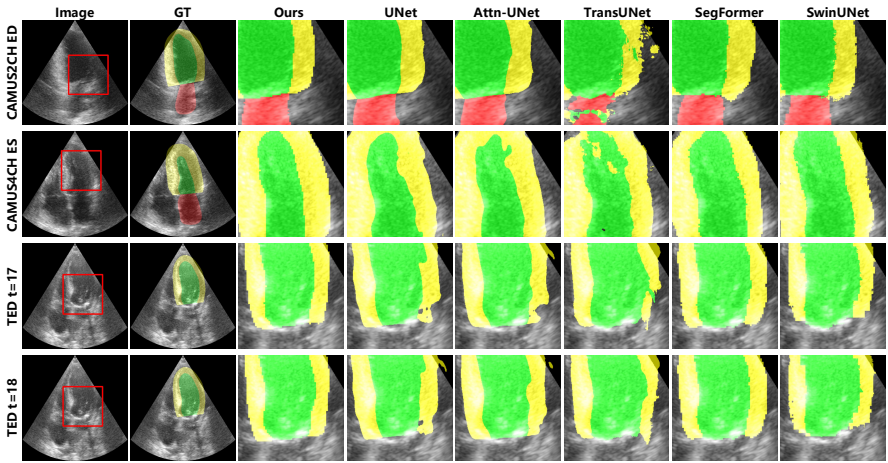}
        \end{center}
	\caption{\textbf{Additional qualitative results on the CAMUS and TED dataset:} We present zoomed-in image patches for detailed comparison. BOTM consistently produces accurate and anatomically coherent segmentations, while baseline methods are adversely affected by various types of noise, leading to anatomically degraded masks.}
    \label{fig:Challenges}
\end{figure*}

\newpage
\definecolor{mygray}{gray}{.92}
\begin{table*}
  \centering
  \scriptsize
  \renewcommand{\arraystretch}{1.1}
  \setlength\tabcolsep{5pt}
  \caption{ Quantitative result breakdowns on CAMUS and TED. For results on CAMUS2CH and CAMUS4CH, we report the average score on [Left] end-diastolic (ED) frame and [Right] end-systolic (ES) frame. For results on TED, we report the average score across all video frames among testset patients.}
  \label{tab:Learning}
\vspace{0.2cm}
  \begin{tabular}{cr||ccccccc}
  \hline
  \rowcolor{mygray}
  &\multicolumn{6}{c}{CAMUS 2CH}\\
  \hline
    \rowcolor{mygray}
  &Methods & mDice~$\uparrow$ & mHD~$\downarrow$  &  mIoU~$\uparrow$ & Spe~$\uparrow$ & MAE~$\downarrow$\\
  \hline
  \multirow{6}{*}{\begin{sideways}LV\end{sideways}} & 
  UNet  & 0.924/0.905 & 9.309/7.187 & 0.881/0.861 & 0.992/0.991  & 0.017/0.013 \\
  &Attn-UNet & 0.922/0.908 & 11.069/12.238 & 0.878/0.866 & 0.992/0.9911  & 0.018/0.013 \\
  &TransUNet & 0.896/0.891 & 10.906/13.593 & 0.841/0.834 & 0.983/0.988  & 0.034/0.031\\
  &SegFormer & 0.907/0.876 &15.657/17.714 &0.832/0.782 &0.992/0.992 &0.020/0.017 \\
  &SwinUNet & 0.918/0.902 & 6.054/9.600 & 0.869/0.878 & 0.990/0.990  & 0.022/0.018 \\
  &\textbf{\textit{BOTM}(Ours)} & \textbf{0.926}/\textbf{0.945} & \textbf{5.841}/\textbf{5.967} & \textbf{0.894}/\textbf{0.893} & \textbf{0.994}/\textbf{0.994}  & \textbf{0.013}/\textbf{0.013} \\
  \hline
  \multirow{6}{*}{\begin{sideways}Myo\end{sideways}} & 
  UNet  & 0.884/0.900 &8.162/6.222 &0.820/0.873 &0.980/0.984 &0.039/0.033 \\
  &Attn-UNet & 0.872/0.904 &10.613/7.535 &0.808/0.879 &0.980/0.984 &0.039/0.032 \\
  &TransUNet & 0.874/0.874 &9.198/11.797 &0.774/0.845 &0.965/0.962 &0.065/0.062\\
  &SegFormer & 0.797/0.809 &23.087/22.575 &0.669/0.684 &0.982/0.983 &0.045/0.041 \\
  &SwinUNet & 0.866/0.878 &9.347/10.877 &0.792/0.831 &0.976/0.981 &0.046/0.040 \\
  &\textbf{\textit{BOTM}(Ours)} & \textbf{0.899}/\textbf{0.937} &\textbf{9.074}/\textbf{6.980} &\textbf{0.873}/\textbf{0.885} &\textbf{0.988}/\textbf{0.980} &\textbf{0.029}/\textbf{0.022} \\
  \hline
  \multirow{6}{*}{\begin{sideways}LA\end{sideways}} & 
  UNet  & 0.870/0.896 &12.866/9.118 &0.866/0.817 &0.996/0.996 &0.011/0.012\\
  &Attn-UNet & 0.887/0.892 &9.403/10.499 &0.830/0.812 &0.996/0.996 &0.011/0.012 \\
  &TransUNet & 0.897/0.860 &9.050/11.550 &0.881/0.632 &0.987/0.987 &0.027/0.027\\
  &SegFormer & 0.830/0.835 &17.480/19.902 &0.722/0.728 &0.997/0.997 &0.013/0.018 \\
  &SwinUNet & 0.884/0.896 &9.488/8.066 &0.867/0.754 &0.992/0.993 &0.015/0.017 \\
  &\textbf{\textit{BOTM}(Ours)} & \textbf{0.901}/\textbf{0.920} &\textbf{7.088}/\textbf{4.636} &\textbf{0.894}/\textbf{0.875} &\textbf{0.998}/\textbf{0.997} &\textbf{0.009}/\textbf{0.009} \\
  \hline
  \rowcolor{mygray}
  &\multicolumn{6}{c}{CAMUS 4CH}\\
  \hline
  \multirow{6}{*}{\begin{sideways}LV\end{sideways}} & 
  UNet  & 0.931/0.901 &7.742/12.273 &0.902/0.859 &0.995/0.993 &0.012/0.011 \\
  &Attn-UNet & 0.918/0.922 &6.214/8.694 &0.902/0.877 &0.994/0.992 &0.012/0.011 \\
  &TransUNet & 0.887/0.867 &12.575/15.851 &0.803/0.727 &0.990/0.985 &0.025/0.023\\
  &SegFormer & 0.929/0.905 &8.579/8.363 &0.869/0.829 &0.994/0.993 &0.015/0.013 \\
  &SwinUNet & 0.918/0.897 &7.111/11.064 &0.852/0.848 &0.988/0.985 &0.018/0.018 \\
  &\textbf{\textit{BOTM}(Ours)} & \textbf{0.949}/\textbf{0.923} &\textbf{4.209}/\textbf{4.035} &\textbf{0.912}/\textbf{0.893} &\textbf{0.998}/\textbf{0.993} &\textbf{0.010}/\textbf{0.011} \\
  \hline
  \multirow{6}{*}{\begin{sideways}Myo\end{sideways}} & 
  UNet  & 0.867/0.861 &11.362/13.606 &0.867/0.860 &0.987/0.988 &0.024/0.024 \\
  &Attn-UNet & 0.865/0.866 &14.419/9.996 &0.864/0.865 &0.987/0.989 &0.024/0.023 \\
  &TransUNet & 0.875/0.849 &18.995/19.510 &0.839/0.717 &0.984/0.983 &0.037/0.038\\
  &SegFormer & 0.831/0.832 &23.986/15.461 &0.715/0.728 &0.985/0.987 &0.029/0.027 \\
  &SwinUNet & 0.888/0.853 &9.174/14.995 &0.874/0.834 &0.982/0.983 &0.033/0.034 \\
  &\textbf{\textit{BOTM}(Ours)} & \textbf{0.909}/\textbf{0.894} &\textbf{8.323}/\textbf{9.912} &\textbf{0.894}/\textbf{0.877} &\textbf{0.990}/\textbf{0.986} &\textbf{0.015}/\textbf{0.025} \\
  \hline
  \multirow{6}{*}{\begin{sideways}LA\end{sideways}} & 
  UNet & 0.876/0.919 &12.477/7.201 &0.850/0.874 &0.996/0.996 &0.009/0.009 \\
  &Attn-UNet & 0.875/0.918 &19.885/10.615 &0.830/0.872 &0.996/0.996 &0.009/0.010 \\
  &TransUNet & 0.816/0.824 &20.997/17.966 &0.727/0.709 &0.988/0.987 &0.027/0.022\\
  &SegFormer & 0.865/0.917 &14.792/8.451 &0.802/0.849 &0.995/0.994 &0.011/0.010 \\
  &SwinUNet & 0.862/0.903 &11.991/9.557 &0.854/0.887 &0.995/0.995 &0.012/0.013 \\
  &\textbf{\textit{BOTM}(Ours)} & \textbf{0.901}/\textbf{0.920} &\textbf{7.435}/\textbf{5.901} &\textbf{0.884}/\textbf{0.895} &\textbf{0.997}/\textbf{0.997} &\textbf{0.009}/\textbf{0.009} \\
  \hline
  \rowcolor{mygray}
  &\multicolumn{6}{c}{TED}\\
  \hline
  \multirow{6}{*}{\begin{sideways}LV\end{sideways}} & 
  UNet  & 0.913 &9.731 &0.878 &0.991 &0.015 \\
  &Attn-UNet & 0.915 &8.825 &0.850 &0.991 &0.016 \\
  &TransUNet & 0.853 &12.646 &0.755 &0.990 &0.027\\
  &SegFormer & 0.901 &11.409 &0.824 &0.991 &0.017 \\
  &SwinUNet & 0.875 &10.532 &0.789 &0.984 &0.022 \\
  &\textbf{\textit{BOTM}(Ours)} & \textbf{0.940} &\textbf{6.387} &\textbf{0.890} &\textbf{0.994} &\textbf{0.012} \\
  \hline
  \multirow{6}{*}{\begin{sideways}Myo\end{sideways}} & 
  UNet  & 0.889 &12.901 &0.811 &0.982 &0.028 \\
  &Attn-UNet & 0.893 &10.314 &0.866 &0.983 &0.028 \\
  &TransUNet & 0.845 &14.609 &0.793 &0.984 &0.036\\
  &SegFormer & 0.871 &17.170 &0.649 &0.984 &0.031 \\
  &SwinUNet & 0.869 &15.767 &0.794 &0.983 &0.037 \\
  &\textbf{\textit{BOTM}(Ours)} & \textbf{0.907} &\textbf{8.652} &\textbf{0.895} &\textbf{0.991} &\textbf{0.016} \\
  \hline
  \end{tabular}
\end{table*}

\end{document}